\documentclass[10pt,twocolumn,letterpaper]{article}

\usepackage[pagenumbers]{cvpr} 

\usepackage{graphicx}
\usepackage{amsmath}
\usepackage{amssymb}
\usepackage{booktabs}
\usepackage{amsthm,amsmath,amssymb}
\usepackage{mathrsfs}
\usepackage{boldline,multirow}

\usepackage{epsfig}
\usepackage{dsfont}

\usepackage{color}

\usepackage{siunitx}
\usepackage{booktabs}
\usepackage{soul}
\usepackage{cite}
\usepackage{cases}
\usepackage{bbding}
\usepackage{url}

\usepackage[pagebackref,breaklinks,colorlinks]{hyperref}
\usepackage{ulem}
\usepackage{dsfont}
\usepackage[capitalize]{cleveref}
\newcommand{\ucite}[1]{\cite{#1}} 
\crefname{section}{Sec.}{Secs.}
\Crefname{section}{Section}{Sections}
\Crefname{table}{Table}{Tables}
\crefname{table}{Tab.}{Tabs.}

\def\cvprPaperID{1758} 
\def\confName{CVPR}
\def\confYear{2023}

\begin{document}

\title{Mixed-order self-paced curriculum learning for universal lesion detection}

\author{Han Li, Hu Han, and S. Kevin Zhou\\
MIRACLE Center, School of Biomedical Engineering \& Suzhou Institute for Advanced Research,\\
University of Science and Technology of China\\
Suzhou 215123, China\\
Key Lab of Intelligent Information Processing of Chinese Academy of Sciences (CAS),\\
Institute of Computing Technology, CAS\\
 Beijing, 100190, China\\
}
\maketitle

\begin{abstract}
Self-paced curriculum learning (SCL) has demonstrated its great potential in computer vision, natural language processing, etc. During training, it implements easy-to-hard sampling based on online estimation of data difficulty. Most SCL methods commonly adopt a loss-based strategy of estimating data difficulty and deweight the `hard' samples in the early training stage. While achieving success in a variety of applications, SCL stills confront two challenges in a medical image analysis task, such as universal lesion detection, featuring insufficient and highly class-imbalanced data: (i) the loss-based difficulty measurer is inaccurate; ii) the hard samples are under-utilized from a deweighting mechanism. To overcome these challenges, in this paper we propose a novel {\bf mixed-order self-paced curriculum learning (Mo-SCL)} method. We integrate both uncertainty and loss to better estimate difficulty online and mix both hard and easy samples in the same mini-batch to appropriately alleviate the problem of under-utilization of hard samples. We provide a theoretical investigation of our method in the context of stochastic gradient descent optimization and extensive experiments based on the DeepLesion benchmark dataset for universal lesion detection (ULD). When applied for two state-of-the-art ULD methods, the proposed mixed-order SCL method can provide a {\bf free boost} to lesion detection accuracy without extra special network designs.
\end{abstract}

\section{Introduction}
\label{sec:intro}
Self-paced curriculum learning (SCL) has recently attracted a lot of attention in computer vision, and natural language processing~\cite{wang2021survey}. The idea behind SCL is to empower vanilla Curriculum Learning (CL) with the ability to automatically estimate the difficulty of data while  following the basic idea of CL: first learning with simple or diverse samples, and then gradually introducing more complex or abstract samples.

Specifically, there are two core components in SCL, namely a data difficulty measurer and a training scheduler. The difficulty measurer usually dynamically estimates a sample difficulty according to its loss, and the training scheduler usually reweights individual  samples' loss in a hard ({\it i.e.}, 0 for hard samples, and 1 for other samples ~\cite{kumar2010self}) or a soft way~\ucite{jiang2014easy, zhao2015self,xu2015multi,gong2018decomposition} to form an easy-to-hard sampling paradigm. This strategy is analogous to the self-learning process of humans.

\begin{figure}[t]
\centering
\setlength{\abovecaptionskip}{0.cm}
\includegraphics[scale=0.55]{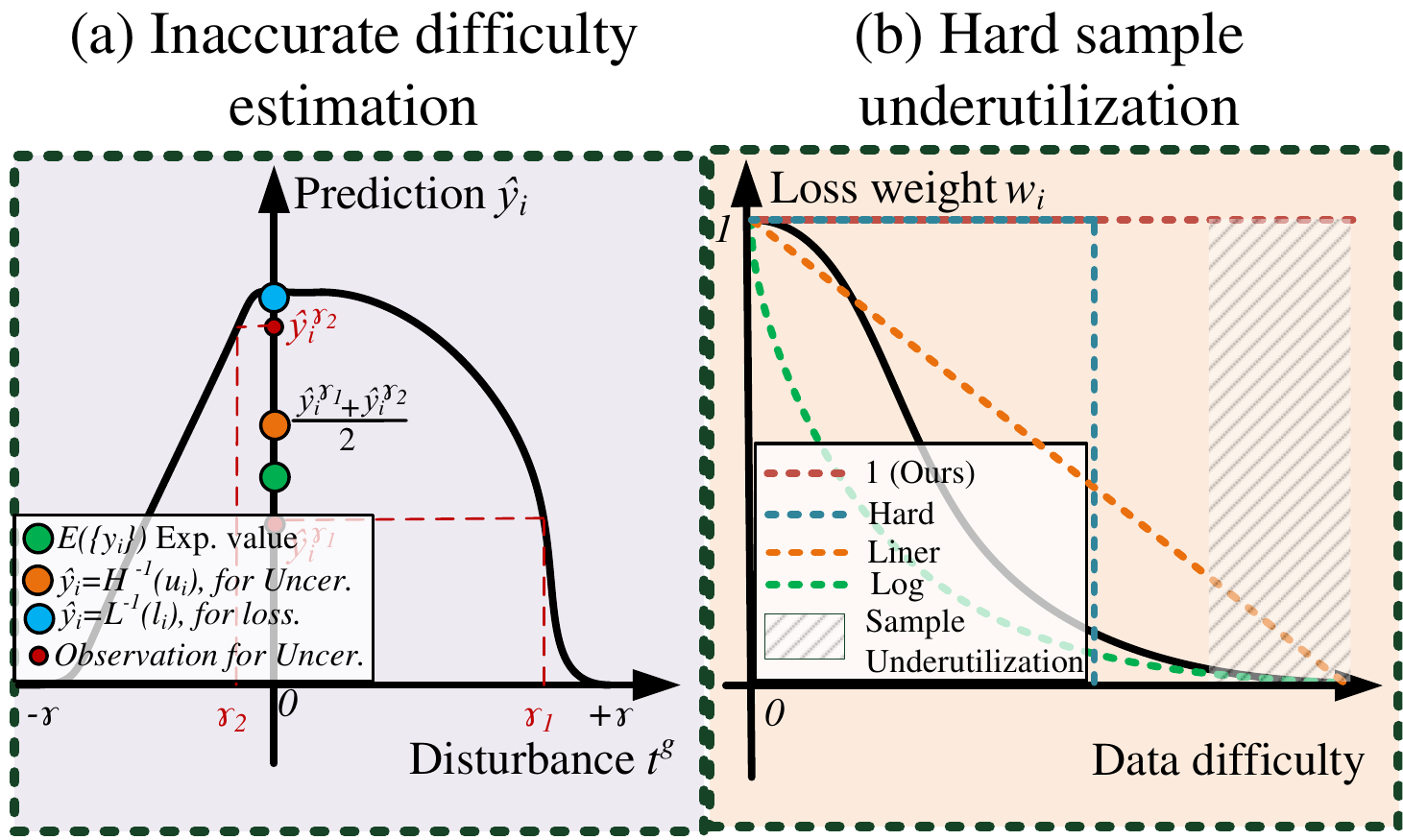}
\caption{(a) The distribution of network prediction $\hat{y}_i$ under different disturbance $t^g$. The green dot is the expected value, which represents the network's robustness. The blue dot is the prediction without $t^g$, which is used to compute the loss $l_i= L(\hat{y}_{i},y_i)$. The yellow dot is the average $\hat{y}_i$ value of the observation red dots, which is used to compute the uncertainty $u_i=H(\frac{\hat{y}_i^{\gamma_1}+\hat{y}_i^{\gamma_2}}{2})$. Obviously, the uncertainty can better estimate the expected value (green). (b) The dashed lines are different loss weighting strategies for samples of different difficulties. Data within the slashed area can easily confront sample under-utilization issue.}
\label{Fig1_overview}
\end{figure}

Despite the encouraging results in natural image analyzing tasks~\ucite{jiang2014easy, zhao2015self,xu2015multi,gong2018decomposition}, existing SCL still confronts two challenges in a medical image analysis task due to the common data deficiency and high-class imbalance issues. (i) {\bf Inaccurate measurer}. The loss-based data difficulty measurer is inaccurate. Intuitively, the loss value computed by a network is a direct and convenient way to measure the sample  difficulty based on the current network status. However, such a measurement using a network's own loss value can be not reliable because of the network's confirmation bias, which will further lead to serious performance degradation. Medical image annotation like annotating lesions in computed tomography (CT) is particularly expensive and time-consuming; therefore a dataset may contain annotations by several different experts or even from different sites. Since the network has a strong fitting capability of all data, it may fit noisy annotations, causing bias in the computed sample loss. If the network's loss cannot accurately reflect individual samples' difficulties, as shown in Fig. \ref{Fig1_overview}, it will further cause the difficult sample not sufficiently explored and the truly difficult samples may not be correctly identified and utilized to improve model generalization ability, leading to (ii) {\bf sample under-utilization}. For example, under the traditional paradigm of SCL using loss for difficulty estimation, CT lesions from small classes or lesions of small sizes can both be determined as hard samples, but the latter ones are more helpful in improving network performance. However, the traditional paradigm of SCL may have under-utilization issue with samples of the latter category.

In this paper, we propose a novel SCL method, termed Mixed-order SCL (Mo-SCL). To cope with the two problems, Mo-SCL introduces both loss and uncertainty for accurately measuring the difficulties of individual samples. In addition, we propose a training scheduler to make use of the scored samples without using the conventional loss reweighting strategy. Specifically, when using loss and uncertainty, instead of using the sum of loss and uncertainty to denote a sample's difficulty, we sort all samples based on loss and uncertainty first and then use the sum of the rank indices to represent a sample's difficulty. Such a method can overcome the limitations of network training fluctuation and non-comparable issues between the loss scale and uncertainty scale. As for the training scheduler, our idea is to simulate human perception behavior, i.e., humans can easily lose concentration if all the samples are of the same difficulty. Therefore, we argue that during network training, the samples in a mini-batch should be mixed in terms of their difficulties. Based on our aforementioned sample difficulties, we can easily  ensure all mini-batch have the almost same difficulty. After that,  we can randomly sample mini-batch to train the network to obtain stable convergence of the network. We evaluate the proposed method for the Universal Lesion Detection
(ULD) task based on two state-of-the-art methods. Our method can consistently improve the performance of both methods  without extra special network designs.

%
%
%
%
%
%
%
%
\section{Related work}
\subsection{Curriculum learning (CL)}
The formal CL concept is first proposed by Bengio {\it et al.} \cite{bengio2009curriculum} with experiments on
supervised visual  and language learning \ucite{liu2022acpl,basu2022surpassing,roy2021curriculum,Huang_2020_CVPR,Pentina_2015_CVPR}. After them, many methods are proposed to pursue generalization improvement
or convergence speedup with the spirit of training from a sequence of easy-to-hard data. To follow this spirit, the  `Difficulty Measurer + Training Scheduler' framework is widely used. Based on the difficulty measurer and training scheduler, CL can be grouped into predefined CL and automatic CL. Specifically, a CL method is defined as predefined CL when both the difficulty measurer and training scheduler are designed by human prior knowledge. If any (or both) of the two
components are learned by data-driven models or algorithms,
then we denote the CL method as automatic CL.

\subsection{Self-paced curriculum learning (SCL)}
Although SCL and CL are usually seen as two different areas, SCL shares the same spirit with CL and also follows the `Difficulty Measurer + Training Scheduler' design. Hence, following \cite{wang2021survey}, we treat SCL as a primary branch of automatic CL which automatically measures a sample's difficulty based on its losses incurred by the current model. Kumar {\it et al.} \cite{kumar2010self} initially propose to deactivate the highly-difficult samples, which is called a hard loss deweighting mechanism. Besides, they are the first to formulate the key principle of SCL model by adding a hard self-paced regularizer (SP regularizer) as shown in Eq.~(\ref{Euq2}) to serve as the training scheduler. As shown in Fig.~\ref{Fig1_overview}, many following works adopt various SP regularizers for better network training, {\it e.g.}, linear \cite{jiang2014easy}, logarithmic \cite{jiang2014easy}, mixture \ucite{jiang2014easy,zhao2015self}, logistic \cite{xu2015multi}, and polynomial \cite{gong2018decomposition}. Albeit with some success, such a loss deweighting mechanism can unavoidably cause sample under-utilization.

Besides, many works also investigate the theoretical  understanding behind SCL \cite{meng2017theoretical} and it also has been widely applied to many practical problems including visual category
discovery \cite{lee2011learning}, segmentation \ucite{kumar2011learning,zhang2017spftn}, image
classification \cite{tang2012self,Ghasedi_2019_CVPR,III_2013_CVPR}, object detection \ucite{tang2012shifting,zhang2019leveraging}, reranking in
multimedia retrieval \cite{jiang2014easy} , person ReID \cite{zhou2018deep}, etc. SCL also works well for pseudo label generation\ucite{jiang2014easy,han2019weakly,ghasedi2019balanced}. Researchers also adopt group-wise weight based on SCL, {\it e.g.}, multi-modal \cite{gong2016multi}, multi-view \cite{xu2015multi}, multi-instance \cite{zhang2015self}, multi-label\cite{li2017self}, multi-task \cite{li2017self2}, etc. Finally,
SCL is used for data-selection-based training strategies, {\it e.g.}, active learning \ucite{lin2017active,tang2019self}.

\subsection{Uncertainty estimation}
Uncertainty estimation has been widely explored in recent years, it can be classified into two categories: Bayesian methods and non-Bayesian methods.
The Bayesian methods model the parameters of a neural network as a posterior distribution using a data sample as input, which provides probability distributions on the output labels~\cite{mackay1992practical}.
Since this posterior distribution is intractable, some approximate variants are proposed for Bayesian methods, {\it e.g.}, Monte Carlo dropout~\cite{gal2016dropout} and Monte Carlo batch normalization~\cite{teye2018bayesian}. For non-Bayesian methods, Deep Ensembles~\cite{lakshminarayanan2016simple} is a representative method that trains multiple models and uses their variance as the uncertainty.

Also, uncertainty estimation methods \ucite{yu2019uncertainty,luo2020deep,xia2020uncertainty,mehrtash2020confidence,shi2021inconsistency} have been used to improve medical image analyzing.
In this paper, we use uncertainty estimation data difficulty.

\subsection{Online hard example mining (OHEM)}
OHEM~\cite{shrivastava2016training} is widely used in many tasks such as segmentation and object detection. The main idea is to select hard examples ({\it e.g.}, triggering a high loss) online and oversample them for the following network training. While the OHEM algorithm has achieved success, 
they easily incur unnecessary wrong information when the training data contains more wrongly or miss annotated samples.

\subsection{Universal lesion detection (ULD)}
ULD in computed tomography (CT), aiming to localize different types of lesions instead of identifying lesion types, plays an essential role in computer-aided  diagnosis (CAD). ULD is clinically valuable but quite challenging because of the diverse lesion shapes, types and expensive annotation, which lead to wrong (or missed) annotation and severe data imbalance. Most existing ULD methods introduce several adjacent 2D CT slices 
as the 3D context information to a 2D detection network  \ucite{zhang2019anchor_free,zhang2020Agg_Fas,yan20183DCE,li2019mvp,yan2019mulan,yang2020alignshift,cai2020deep,zhang2020revisiting,tang2021weakly,yang2021asymmetric,li2021conditional,lyu2021segmentation} or directly adopt 3D designs \cite{cai2020deep}.

\begin{figure*}
\centering
\includegraphics[scale=0.75]{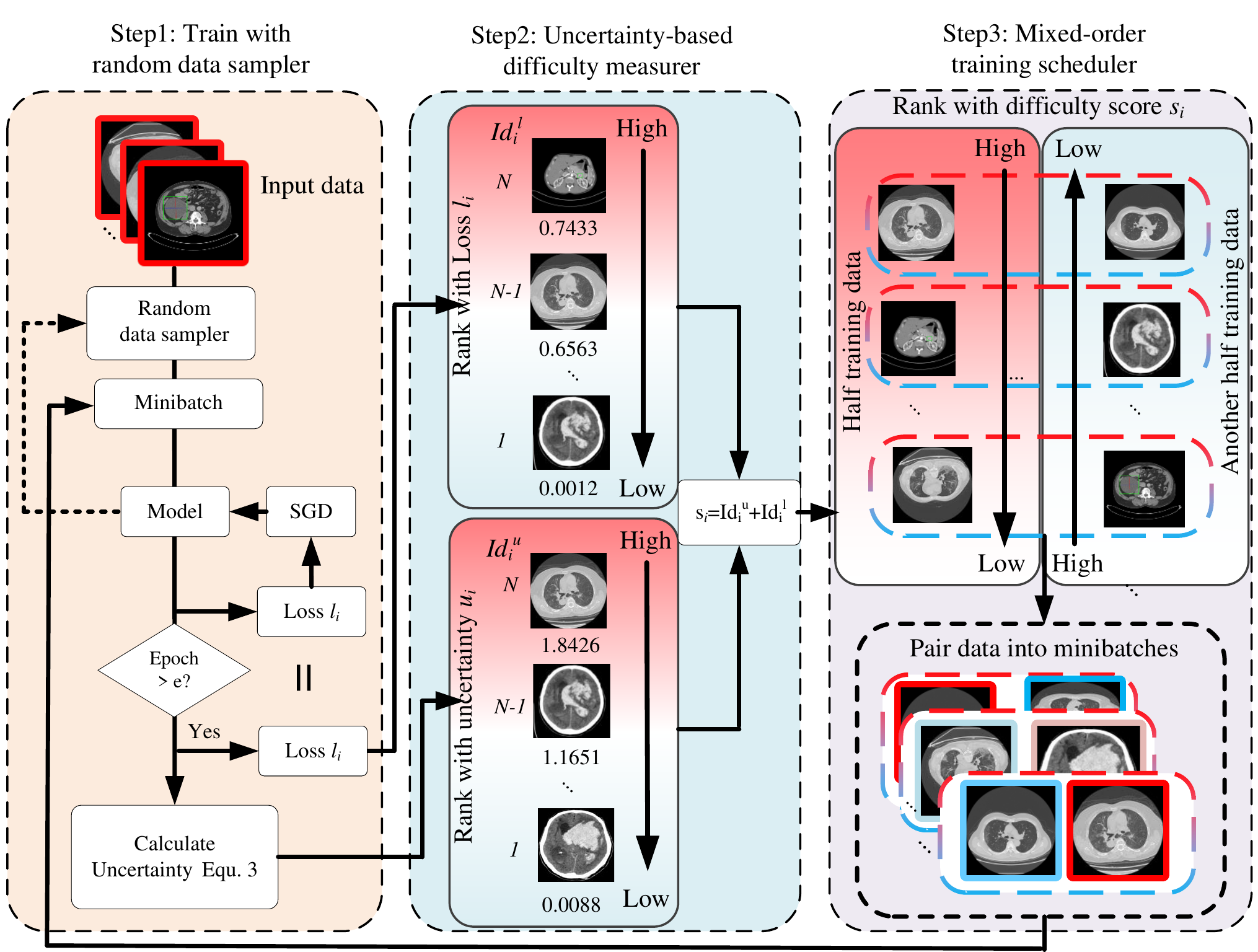}
\caption{There are three steps of our method: Step-1: For the first training stage, we use the vanilla random sampler to sample data for network training, after $e$ epochs we activate the uncertainty estimation component (Equ. \ref{Equ3}) to calculate each sample's uncertainty. Step-2: we sort all samples based on loss and uncertainty first, and then use the sum of the rank indices $\mathcal{X}_u$ and $\mathcal{X}_l$ as its difficulty score $s$. Step-3: we first rank all training samples according to the difficulty score $s$ and then pair samples with high $s$ with that with low $s$ in a mini-batch. The newly formed mini-batches are used for the following network training.}  \label{fig2_network_architecture}
\end{figure*}

\section{Proposed method}
In this section, we first introduce the conventional  SCL method in \ref{sec.scl}, then detail the proposed Mo-SCL components in \ref{sec.measurer} and \ref{sec.motrain}. We also provide theoretical proofs to support the superiority of the proposed two designs for difficulty measurer and mixed-order scheduler.

\subsection{Self-paced curriculum learning} \label{sec.scl}
Let $\mathcal{X}=\{(x_i,y_i)\}_{i=1}^{I}$ represent the training data, where $x_i$ is the $i-th$ input data and $y_i$ is its corresponding label. Ideally, we want to find a model $\hat{y_i}=\mathcal{F}(x_i|\theta)$ with parameter $\theta$ to approximate the label $y_i$. The difference between   $\hat{y_i}$ and  $y_i$ is measured by a risk function (or loss function). To minimize the risk function, SGD is a popular optimization method. The vanilla SGD updates the model weight $\theta$ iteratively based on mini-batch $\mathcal{B}$, sampled from
the training data based on some probability function $\mathcal{D}(\mathcal{X})$.

SCL possesses two components of data difficulty measurer and training scheduler. The data difficulty measurer in SCL simply uses the training loss $l_i=L(\hat{y_i},y_i)$ with respect to $\mathcal{F}(\cdot|\theta)$ as data $x_i$'s difficulty score, where $L(\cdot)$ is the predefined loss function (or risk function). The training scheduler in SCL is to reweight samples loss according to their loss with weight $v^{*}$, the $v^{*}$ function follows some predefined rules like:
\begin{equation}
  v^{*}(l,\lambda)=\left\{
  \begin{aligned}
 1&~~~~if~l<\lambda \\
0&~~~~else\\
  \end{aligned}
  \right . \label{Euq1}
\end{equation}
where $\lambda$ is a hyperparameter called age. Formally, \cite{kumar2010self} involved a self-paced regularizer $f(v,\lambda)$ to formulate SCL as follows:
\begin{equation}
v^{*}(l,\lambda)= \underset{v\in [0,1]}{\arg\max} \, [vl+f(v,\lambda)]. \label{Euq2}
\end{equation}
Then if $v^{*}(l,\lambda)$ follows Eq.~(\ref{Euq1}), it is easy to derive~\cite{kumar2010self} that $f(v,\lambda)=-\lambda v$. 

\subsection{Uncertainty- \& loss-based difficulty measurer} \label{sec.measurer}
In this part, we first introduce our difficulty measurer, then we try to prove that our difficulty measurer is a distribution-based method while the vanilla loss-based measurer is point-based one.

\noindent{\bf Uncertainty estimation.} As shown in Fig. \ref{fig2_network_architecture}(step 1), the uncertainty-estimating process is activated after a certain epoch $e$ (converged). For an image $x_i$ and current model $\mathcal{F}(\cdot|\theta)$, the uncertainty $u_i$ is computed as the information entropy of the average prediction of the model $\hat{y_i}=\mathcal{F}(x_i|\theta)$ under $G$ different disturbances $\{t^{1},t^{2},...,t^{G}\}$:

\begin{equation}
\begin{aligned}
   \hat{y}_{i}^{g}=\mathcal{F}(x_i|\theta, t^{g} ),~~
   u_i=H(\frac{1}{|G|}\sum_{g=1}^{G} \hat{y}_{i}^{g}),
 \end{aligned} \label{Equ3}
  \end{equation}
where $H(p) \triangleq -p \times log(p)$ is the information entropy.
$\hat{y}_{i}^{g}$ means the prediction of  $x_i$ via network $\mathcal{F}(\cdot)$ under disturbance $t^{g}$.
   It is worth noting that we directly add noise $t^{g}$ to key feature maps $f$ ({\it e.g.}, the encoder's output of segmentation networks, backbone feature maps of two-stage detection methods) instead of to the input images. This is because noises to input images  can not obviously disturb the network's output ~\cite{ouali2020semi}.
   The feature map $\hat{f^{g}}(t^{g})$ under disturbance $t^{g}$ is formulated as:
    \begin{equation}
      \begin{aligned}
          \hat{f}^{g}(t^{g})= f \otimes (\mathds{1}+t^{g}),
      \end{aligned}
    \end{equation}
   where $\otimes$ is the pixel-level dot multiplication, $\mathds{1}$ is a matrix with the same size as \textit{f} and filled with the scalar value $1$.
   Each pixel value of $t^{g}$ is sampled from a uniform distribution $U[-\gamma,+\gamma]$.

\noindent{\bf Uncertainty- and loss-based ranking.} As shown in Fig. \ref{fig2_network_architecture} (step 2), we rank all training samples based on their uncertainty and loss values, respectively.

\begin{equation}
  \begin{aligned}
      \mathcal{X}_u= [x_{(\cdot,j)}, y_{(\cdot,j})],~~~~s.t.~~u{(\cdot,j)}\geq u{(\cdot,j+1)} \\
      \mathcal{X}_l= [x_{(\cdot,k)}, y_{(\cdot,k)}],~~~~s.t.~~l{(\cdot,k)}\geq l{(\cdot,k+1)}
  \end{aligned}
\end{equation}
For the same training sample, we use the sum of its index in $\mathcal{X}_u$ and $\mathcal{X}_l$ as its \underline{difficulty score $d$}. Such a method can overcome the limitations of network training fluctuation and non-comparable issues between the loss scale and uncertainty scale.

\noindent{\bf The superiority of introducing uncertainty.}
In this part, we provide theoretical proof to support the superiority of introducing uncertainty. To prove it: (i) the first target is to prove the necessity of introducing uncertainty. Obviously, it is necessary only if uncertainty is also a difficulty estimation and it focuses on different dimensions for difficulty estimation compare with loss-based measure; (ii) then the second target (if possible) is to prove that uncertainty is superior to loss for difficult measurement. Considering the above two targets, we introduce:

\textbf{Lemma 1.} For training sample $x_i$ and model $\mathcal{F}(x_i|\theta)$, the uncertainty $u_i$ is the uncertainty estimation of expected value $\mathbb{E}(\hat{Y}_{i}^{g})$ of distribution $\hat{Y}_{i}^{G}=[\hat{y}_{i}^{g}]$, and loss $l_i$ is just one point in this distribution which is mapped from $\hat{y}_i$.

It is obvious that when Lemma1 is true, uncertainty is a distribution-based difficulty estimation that is different from the point-based loss (the first target satisfied). Also, statistically, the distribution-based estimation should be more accurate than a point-based  one (the second target satisfied).

\underline{\bf Proof.}
Obviously, when we set disturbance $t_{g}=0$, we can get:
\begin{equation}
  \begin{aligned}
    \hat{y}_i \equiv \hat{y}_{i}^{g}~~s.t.~~t_{g}=0,\\
    [t_{g}=0] \in U[-\gamma,+\gamma].
  \end{aligned}
\end{equation}
Then the loss for this sample $x_i$ can be formulated as:
\begin{equation}
  \begin{aligned}
    l_i=L(y_i,\hat{y}_i) \equiv L(y_i,\hat{y}_{i}^{g})~~s.t.~~t_{g}=0,
  \end{aligned}
\end{equation}
It is easy to prove that for conventional  loss functions, their inverse functions exist (please refer to supplemental for more details). For example, Cross entropy (CE) is monotonous in $\hat{y}_{i}^{g}\in [0,1]$ and L2 Loss is monotonous with respect to $|y_i-\hat{y}_{i}^{g}| \in [0,1]$, thus we can get the inverse functions of these losses:
\begin{equation}
  \begin{aligned}
    \hat{y}_{i}^{g}=L^{-1}(y_i,l_i)~~s.t.~~t_{g}=0,
  \end{aligned}
\end{equation}
Also, we can prove that the information entropy function also has an inverse function with respect to $y_i \in [0,1]$ (refer to supplemental for more details), revisit the uncertainty value $u_i$ definition in \ref{Equ3}, we can get:
\begin{equation}
  \begin{aligned}
    lu_i&= H(\hat{y}_{i}^{g})=H(L^{-1}(y_i,l_i))~~s.t.~~t_{g}=0,\\
    u_i&=H(\frac{1}{|G|}\sum_{g=1}^{G} \hat{y}_{i}^{g})\equiv H(\mathbb{E}[\sum_{g=1}^{G} \hat{y}_{i}^{g})],\\
    l_i&=L(y_i,\hat{y}_{i}^{g})=L(y_i,H^{-1}(lu_i))~~s.t.~~t_{g}=0,
  \end{aligned}\label{Equ9}
\end{equation}
where $lu_i$ is the loss-based uncertainty score. This concludes Lemma 1.

\textbf{Corollary 1.}
Statistically, the distribution-based estimation uncertainty $u_i$ is more accurate in reflecting data difficulty than point-based loss $l_i$ or $\hat{y}_i$ does.

Besides, in the subsequent Corollary 4.1, we further investigate the superiority of uncertainty from the aspect of network robustness.

\subsection{Mixed-order training scheduler} \label{sec.motrain}
After we get the difficulty score $d$ for each training sample, we imitate the fact that humans easily lose their concentration and interest if all learning materials are relative easy (or hard) and build mini-batches for network training in a new way.

As shown in Fig. \ref{fig2_network_architecture} (step 3), we pair the samples with high $d$ with those with low $d$ in a mini-batch, we set batch size $b=2$ for convenience in explanation:
\begin{equation}
B= [x_m,x_n],~~~s.t.~~m \neq n,~~ m,n\in \{1,2,\ldots,N\},
\end{equation}
where $B$ is the training mini-batches. The newly formed mini-batches are used for the following network training.


Four typical situations of data can be intuitively  defined according to data loss and uncertainty:
\begin{enumerate}
  \item[$\bullet$] {Data with high uncertainty $u^h$ and high loss $l^h$: data class which has insufficient  samples or conflicts with the majority-class data.}
  \item[$\bullet$]{Data with low uncertainty $u^l$ and high loss $l^h$: data class which is wrongly labeled or miss labeled.}
  \item[$\bullet$]{Data with low uncertainty $u^l$ and low loss $l^l$: majority-class data or data with enough training samples.}
  \item[$\bullet$]{Data with high uncertainty $u^h$ and low loss $l^l$: Noisy data or overfitted samples by the network.}
\end{enumerate}
where $u^h$ and $u^l$ (or $l^h$ and $l^l$) denote the uncertainty (or loss) is relative high and low.

As will be shown later, keeping the mini-batches' total difficulty score stable among the  whole training dataset can be helpful to avoid unnecessary data conflicts and relieve sample under-utilization (or improve the effectiveness of hard-sample learning). Hence, we can analyze the following situations:
\begin{enumerate}
  \item[$\bullet$] {\{$<u^h,l^h>$,$<u^l,l^l>$\}: the latter gradient causes a minor effect on the former ones, thus the former is the main contributor of the gradient.}
  \item[$\bullet$]{\{$<u^l,l^h>$,$<u^h,l^l>$\}: both influences each other, but cause less effect on
$<u^h,l^h>$ or $<u^l, l^l>$ data.}
\end{enumerate}
where the \{$<u^h,l^h>$,$<u^l,l^l>$\} can be treated as an example that relieves sample under-utilization (or improves the effectiveness of hard-sample learning), while \{$<u^l,l^h>$,$<u^h,l^l>$\}  for avoiding unnecessary data conflicts.



\noindent\textbf{The superiority of mixed-order training scheduler.}
In this part, we first show the reasonableness of the abovementioned two assumptions, i.e., avoid unnecessary data conflicts and relieve sample under-utilization. We predefined some annotations  for easy descriptions.

Our proof is based on sigmoid-based (or softmax-based) network $h(x)=\sigma(x_i^{t}w)=\sigma(z)$, where $w$ is the network weight and $\sigma$ is the sigmoid function for binary classification tasks (or softmax for multi-class tasks), and $z$ is the latent feature input of the sigmoid (or softmax) function.

For a better explanation, we define the loss function as the mean square error (MSE) loss:
\begin{equation}
  \begin{aligned}
    L(h(x_i|w),y_i)=&(h(x_i|w)-y_i)^2=(x_{i}^{t}w-y_i)^2\\
    =&L(\mathcal{X}_{i},w)
  \end{aligned}
\end{equation}
But other popular loss functions still hold the following conclusions in this work.
We assume that the $w_t$ represents the converged weight with a random sampling manner over the whole training dataset. We use the SGD optimizer to optimize $w$:
\begin{equation}
    w_{t+1}= w_t - \eta s; ~~  s=\frac{ \partial L(\mathcal{X}_i,w_t) }{ \partial w_t},
\end{equation}
where $\eta$ is the learning rate value, $s$ is the gradient of $\mathcal{X}_i$ at time $t$.

Now, we go back to the target, avoiding unnecessary data conflicts and relieve sample under-utilization. We can find that both two assumptions are data conflict problems, the unnecessary data conflicts means wrongly/miss annotated data ( $<u^l,l^h>$) or noisy samples ($<u^h,l^l>$) causing  effect on easy clean samples ($<u^l, l^l>$) or insufficient data ($<u^h,l^h>$), relieving sample under-utilization means pairing easy clean samples ($<u^l, l^l>$) with the insufficient data ($<u^h,l^h>$) yield minor gradient conflict on the insufficient data.

 The data conflict is mostly equivalent to gradient conflicts in gradient-descent-based methods. Such gradient conflicts are hard to observe or measure;  thus we need an explicit score like losses to measure it.

\textbf{Lemma 2.}
For a converged model trained based on random sampling, the gradient conflict (or gradient intersection angle) with respect to $\hat{y_i}$ between two samples monotonically increases with respect to their total loss.

If Lemma 2 is true, we can simply use losses to estimate the conflict, and thus adopt mini-batches like \{$<l^h>$,$<l^l>$\} to avoid unnecessary data conflicts and relieve sample under-utilization simultaneously.

\underline{\bf Proof.} By contradiction. Assume that, for a converged model trained based on random sampling, the gradient conflicts (or gradient intersection angle) between two samples monotonically decrease with respect to their total loss.
Suppose we have three training samples $[x_1,x_2,x_3]$ where $l_1 \textless l_2 \textless l_3$ for all time $t$. As we set a mini-batch size as 2, a training sample is oversampled to form  $[x_1,x_2,x_3,x_2]$. Based on the random sampling scheduler, we have four mini-batch combinations:
\begin{equation}
  \begin{aligned}
    B_1=[x_1,x_2],B_2=[x_2,x_3]\\
    B_3=[x_1,x_3],B_4=[x_2,x_2]
  \end{aligned}
\end{equation}
Thus we get two gradient steps for $x_1$ and $x_3$ respectively:
\begin{equation}
  \begin{aligned}
    |s_{(1,y)}|=& |s_{(1,y)}^{t1}|+|cos\theta_{(1,2)}s_{(2,y)}^{t1}|+|s_{(1,y)}^{t3}|+cos\theta_{(1,3)}|s_{(3,y)}^{t3}|\\
    |s_{(3,y)}|=& |s_{(3,y)}^{t2}|+|cos\theta_{(3,2)}s_{(2,y)}^{t2}|+|s_{(3,y)}^{t3}|+cos\theta_{(3,1)}|s_{(1,y)}^{t3}|\\
    s_{(\cdot,y)}=&\frac{ \partial L(\hat{y_i},y_i) }{ \partial \hat{y_i}}=2(\hat{y_i}-y_i)
  \end{aligned}\label{Equ14}
\end{equation}
where $s_{(\cdot,y)}$ is the loss gradient respect to $\hat{y_i}$, $\theta_{(1,2)}$ is the angle between $s_1$ and $s_2$ Then we compare two gradient steps:

\begin{equation}
  \begin{aligned}
    \Delta(|s_{(\cdot,y)}|) =&|s_{(1,y)}| - |s_{(3,y)}|\\
     =& |s_{(1,y)}^{t1}|-  |s_{(3,y)}| +|s_{(1,y)}^{t3}|-|s_{(3,y)}^{t3}|\\
    +&cos\theta_{(1,2)}|s_{(2,y)}^{t1}| - cos\theta_{(3,2)}|s_{(2,y)}^{t2}|
  \end{aligned}
\end{equation}
For we assume that for any time $t$ $l_1 \textless l_3$, we can easily reach that $|s_{(1,y)}^{t1}|-  |s_{(3,y)}| +|s_{(1,y)}^{t3}|-|s_{(3,y)}^{t3}| \textless 0$ based on Eq.~(\ref{Equ14}). Also,  $cos\theta_{(1,2)}|s_{(2,y)}^{t1}| - cos\theta_{(3,2)}|s_{(2,y)}^{t2}| \textless 0$ for the gradient conflicts between $s_1$ and $s_2$ is greater than that between $s_3$ and $s_2$. Hence, $\Delta(s_{(\cdot,y)}) \textless 0$. Now we can easily find that after enough SGD iterations, $l_3 \textless l_2$, which conflicts with the assumption.

\textbf{Corollary 2.1}
Loss can be used for gradient conflict estimation with respect to $\hat{y_i}$.

\textbf{Corollary 2.2}
Use mini-batch= \{$<l^h>$,$<l^l>$\} to avoid unnecessary data conflicts and relieve sample under-utilization simultaneously, with respect to $\hat{y_i}$.

\textbf{Corollary 2.3}
Pairing samples with more gradient conflicts into one minibatch can cause low-efficiency and unstable training with respect to $\hat{y_i}$.

Obviously, Corollaries 2.1-2.2 do not accord with the experiment observation, this is because the theory that $s_{(\cdot,y)}=\frac{ \partial L(\hat{y_i},y_i) }{ \partial \hat{y_i}}=2(\hat{y_i}-y_i)$ only satisfied with respect to $\hat{y_i}$. In other words, they are true only if no nonlinear operation in networks. Hence, with respect to $z$ (latent feature before sigmoid), we should first recalculate the gradient step.

\textbf{Lemma 3.}
The gradient scales on latent feature $z$ of sigmoid-based models are related to the prediction $\hat{y_i}$ and risk $y_i -\hat{y_i}$.

\underline{\bf Proof.}
If the ground truth is $y_i=1$,
\begin{equation}
  \begin{aligned}
    s_{(\cdot,z)}=&{\frac{ \partial L(\hat{y_i},y_i) }{ \partial \hat{y_i}}} \cdot {\frac{ \partial \hat{y_i} }{ \partial z_i}}=2(\hat{y_i}-y_i)[(\hat{y_i})(1-\hat{y_i})]\\
    =&-2\hat{y_i}(1-\hat{y_i})^2= - 2\hat{y_i} l_i
  \end{aligned}
\end{equation}
If the ground truth is $y_i=0$, $s_{(\cdot,z)}=2\hat{y_i}^2(1-\hat{y_i})$.
With this loss function, the gradient scale on $z$ of sigmoid-based models is related to the prediction $\hat{y_i}$ and risk $y_i -\hat{y_i}$. It is also easy to prove that other common losses and multi-class tasks still follow this rule.

\textbf{Corollary 3.1}
Now we can rewrite Lemma 2:  For a converged model trained based on random sampling, the gradient conflict (or gradient intersection angle) with respect to $z$ between two samples monotonically increases with respect to $|s_{(\cdot,z)}|=2 \hat{y_i} l_i$ if $y_i=1$ or $|s_{(\cdot,z)}|=2 \hat{y_i}^2 l_i^{1/2}$ if $y_i=0$.

Now it is clear that the gradient conflict on latent feature $z$ is related to the prediction $\hat{y_i}$ and loss $y_i -\hat{y_i}$. Recall that uncertainty is just a distribution-based version of losses in Lemma 1.

\textbf{Corollary 3.2}
The uncertainty $u_i$ is the uncertainty estimation of expected value $\mathbb{E}(\hat{Y}_{i}^{g})$ of distribution $\hat{Y}_{i}^{G}=[\hat{y}_{i}^{g}]$, while loss $l_i$ is one point in this distribution which is mapped from $\hat{y_i}$. The uncertainty $u_i$ should be capable of representing the confidence of $l_i$ and $\hat{y_i}$.
The uncertainty $u_i$ can be used to guide $s_{(\cdot,z)}$ and reflect the data difficulty.

Now we reconsider what the uncertainty essentially is.

\textbf{Lemma 4.}
Uncertainty is a reflection of the robustness of the network.

\underline{\bf Proof.}
First, we give out the definition of network robustness. For binary classification tasks, let $Y_i({\hat y}_i)$ denotes the output label respect to $y_i$:
\begin{equation}
  Y_i({\hat y}_i)=\left\{
  \begin{aligned}
 1&~~~~if~{\hat y}_i>0.5 \\
0&~~~~else\\
  \end{aligned}
  \right .
\end{equation}
When adding reasonable a perturbation to the network ({\it e.g.}, inputs, architecture, feature), the ability to keep the classification result $Y_i({\hat y}_i)$ stable is called robustness.

Recall the definition of uncertainty in Eqs.~(\ref{Equ3}) and (\ref{Equ9}):
\begin{equation}
\begin{aligned}
   \hat{y}_{i}^{g}=\mathcal{F}(x_i|\theta, t^{g} ),~~~
   u_i=H(\frac{1}{|G|}\sum_{g=1}^{G} \hat{y}_{i}^{g}),\\
   \hat{f}^{g}(t^{g})= f \otimes (\mathds{1}+t^{g}),
 \end{aligned} \label{equ:socre}
  \end{equation}
The uncertainty $u$ is computed as the information entropy of the average prediction of model $\hat{y_i}=\mathcal{F}(x_i|\theta)$  under $G$ different disturbances $t^{1},t^{2},...,t^{G}$.
Each pixel value of $t^{g}$ is sampled from a uniform distribution $U[-\gamma,+\gamma]$. Thus, it can reflect the robustness of the network.

\textbf{Corollary 4.1}
Now based on Lemma 4 and Corollary 3.1, we can prove the intuitive understanding, that is, keeping mini-batches' total difficulty score even along the whole training dataset is helpful for avoiding unnecessary data conflicts thus leading network robust training.

\begin{figure}[t]
\centering
\setlength{\abovecaptionskip}{0.cm}
\includegraphics[scale=0.55]{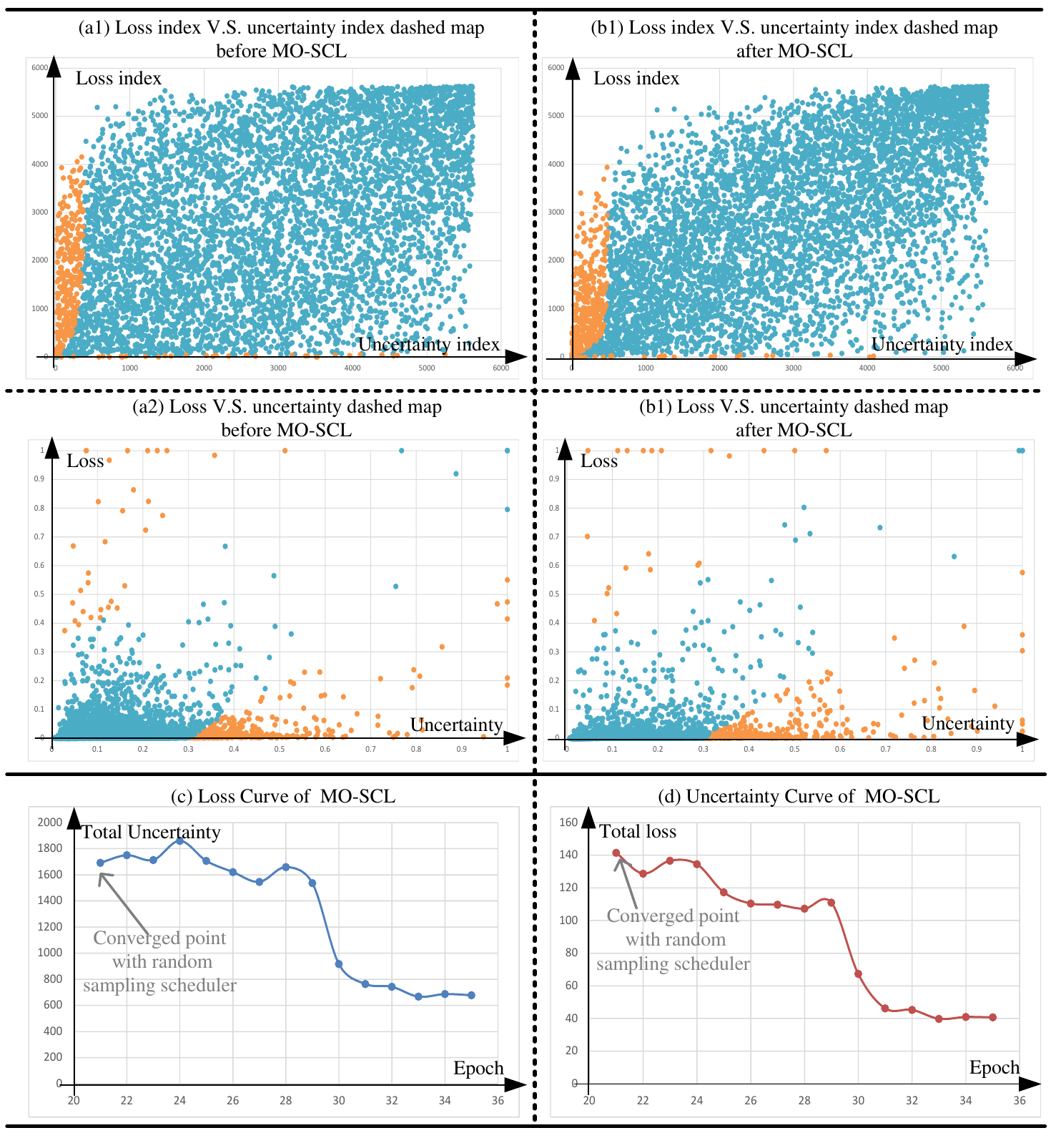}
\caption{Illustration of loss and uncertainty relationship based on SATr \cite{li2022satr}. Yellow (or cyan) points denote the sample whose absolute difference between uncertainty and loss is greater (or less than) than 0.3.}\label{Fig3_loss_uncertainty}

\end{figure}
\newsavebox{\tablebox}
\begin{table*}[ht]
\centering
\caption{Sensitivity ($\%$) at various FPPI on the testing dataset of DeepLesion \cite{yan18deeplesion} under 25$\%$  and 50 $\%$ training data settings. Mo+l, Mo+u, Mo+l+u denote using loss-base, uncertainty-based, and loss+uncertainty-based difficulty measurers with our Mixed-order scheduler. $\hat{Mo}$ denotes anti mixed-order data pairing which pairs high- (or low-) difficulty data with high- (or low-) difficulty data into one mini-batch, i.e, hi+hi.} \label{results}
\begin{lrbox}{\tablebox}
\begin{tabular}{p{35mm}p{13mm}<{\centering}p{12mm}<{\centering}p{12mm}<{\centering}p{12mm}<{\centering}p{23mm}<{\centering}p{23mm}p{23mm}p{23mm}p{23mm}p{23mm}}

  \toprule[1.5pt]
  \textbf{Methods}&\textbf{Data}&\textbf{Difficulty}&\textbf{Scheduler}&\textbf{Deweight}&\textbf{$@0.5$}&\textbf{$@1$}&\textbf{$@2$}&\textbf{$@4$}&Avg.[0.5,1,2,4]\\
\toprule[1pt]
A3D\cite{yang2021asymmetric}&25\% &-& Random &-&55.67&65.39 &73.35& 79.31&68.43\\
A3D+loss deweight&25\% &l& Random &Liner&54.28 (1.39$\downarrow$)&63.99 (1.40$\downarrow$) &72.18 (1.17$\downarrow$)& 78.97 (0.34$\downarrow$)&(3.82$\downarrow$)\\

A3D+$\hat{Mo}$+l+u&25\% &l+u& hi+hi &-&56.90 (1.73$\uparrow$)&66.19 (1.80$\uparrow$) &74.63 (1.28$\uparrow$)& 80.21 (0.90$\uparrow$)&(1.82$\downarrow$)\\

A3D+Mo+l+u&25\% &l+u& hi+low &-&60.34 (4.67$\uparrow$)&69.38(3.99$\uparrow$) &75.28 (1.95$\uparrow$)& 82.45 (3.14$\uparrow$)&71.86(3.43$\uparrow$)\\

\midrule
SATr\cite{li2022satr}&25\% &-& Random &-&59.99 &68.05 &74.67 & 79.09 &70.45\\

SATr+loss deweight&25\% &l& Random &Liner&58.17 (1.82$\downarrow$)&67.45 (0.60$\downarrow$) &73.84 (0.83$\downarrow$)& 78.29 (0.80$\downarrow$)&69.44(1.01$\downarrow$)\\

SATr+$\hat{Mo}$+l+u&25\% &l+u& hi+hi &-&65.61 (5.62$\uparrow$)&72.50 (4.45$\uparrow$) &77.87 (3.20$\uparrow$)& 81.82 (2.73$\uparrow$)&74.45(4.00$\uparrow$)\\

SATr+Mo+l&25\% &l& hi+low &-&65.17 (5.18$\uparrow$)&71.88 (3.83$\uparrow$) &77.30 (2.63$\uparrow$)& 81.93 (2.84$\uparrow$)&74.07(3.62$\uparrow$)\\

SATr+Mo+u&25\% &u& hi+low &-&66.54 (6.65$\uparrow$)&73.87(5.82$\uparrow$) &79.24 (4.57$\uparrow$)& 84.25 (5.16$\uparrow$)&75.98(5.53$\uparrow$)\\

SATr+Mo+l+u&25\% &l+u& hi+low &-&\textbf{68.54} (8.55$\uparrow$)&\textbf{75.38} (7.33$\uparrow$) &\textbf{80.64} (5.97$\uparrow$)& \textbf{84.83} (5.74$\uparrow$)&\textbf{77.35}(6.90$\uparrow$)\\

\bottomrule[1.5pt]

A3D\cite{yang2021asymmetric} &50\% &-& Random &-&72.52&80.27&86.14&90.15&82.27\\
A3D+loss deweight &50\% &l& Random &Liner&70.85 (1.67$\downarrow$)&78.80 (1.47$\downarrow$) &85.12 (1.02$\downarrow$)& 89.73(0.42$\downarrow$)&81.13(1.14$\downarrow$)\\

A3D+$\hat{Mo}$+l+u&50\% &l+u& hi+hi &-&71.87 (0.65$\downarrow$)&79.45 (0.82$\downarrow$) &85.60 (0.54$\downarrow$)& 89.30 (0.85$\downarrow$)&81.56(0.71$\downarrow$)\\

A3D+Mo+l+u&50\% &l+u& hi+low &-&74.00 (1.48$\uparrow$)&81.23 (0.96$\uparrow$) &86.48 (0.34$\uparrow$)& 90.11 (0.04$\downarrow$)&82.96(0.69$\uparrow$)\\

\midrule
SATr \cite{li2022satr} &50\% &-& Random &-&75.24&82.19 &86.99&90.96&83.85\\
SATr+loss deweight &50\% &l& Random &Liner&74.63 (0.61$\downarrow$)&81.43 (0.76$\downarrow$) &86.18 (0.81$\downarrow$)& 90.00 (0.96$\downarrow$)&83.06(0.79$\downarrow$)\\

SATr+$\hat{Mo}$+l+u&50\% &l+u& hi+hi &-&74.52 (0.72$\downarrow$)&81.83 (0.36$\downarrow$) &86.69 (0.30$\downarrow$)& 90.63 (0.33$\downarrow$)&83.42(0.43$\downarrow$)\\

SATr+Mo+l&50\% &l& hi+low &-&73.36 (1.88$\downarrow$)&80.52 (1.67$\downarrow$) &85.40 (1.59$\downarrow$)& 89.14 (1.82$\downarrow$)&82.11(1.74$\downarrow$)\\

SATr+Mo+u&50\% &u& hi+low &-&75.69 (0.45$\uparrow$)&82.55 (0.36$\uparrow$) &87.12 (0.13$\uparrow$)& 90.99 (0.03$\uparrow$)&84.09(0.24$\uparrow$)\\

SATr+Mo+l+u&50\% &l+u& hi+low &-&\textbf{76.97} (1.73$\uparrow$)&\textbf{83.66} (1.47$\uparrow$) &\textbf{87.27} (0.28$\uparrow$)& \textbf{91.11} (0.15$\uparrow$)&\textbf{84.75(}0.90$\uparrow$)\\

\bottomrule[1.5pt]
A3D\cite{yang2021asymmetric} w/ GT ROI &50\% &-& Random &-&93.45&95.63&97.22&98.39&96.17\\
SATr \cite{li2022satr} w/ GT ROI &50\% &-& Random &-&94.04&96.00 &97.30&98.57&96.48\\
\bottomrule[1.5pt]
\end{tabular}

\end{lrbox}
\scalebox{0.63}{\usebox{\tablebox}}
\end{table*}

\section{Experiment}
\subsection{Dataset and setting}
Our experiments are conducted on the ULD dataset DeepLesion\cite{yan18deeplesion}.  The dataset contains 32,735 lesions on 32,120 axial slices from 10,594 CT studies of 4,427 unique patients. Most existing datasets typically focus on one type of lesion, while DeepLesion contains a variety of lesions with a large diameter range (from 0.21 to 342.5mm). The 12-bit intensity CT is rescaled to [0,255] with different window range settings used in different frameworks. Also, every CT slice is resized and interpolated according to the detection frameworks' settings. We follow the official split, i.e., $70\%$ for training, $15\%$ for validation and $15\%$ for testing. To further test our method's performance on a small dataset, we also conduct experiments based on  25\% and 50\% training data.  The number of false positives per image (FPPI) is used as the evaluation metric. For training, we use the original network architecture and settings. As for the loss selection, we use the anchor classification loss in Region Proposal Network (RPN) for data difficulty measurement. As for the uncertainty calculation, we add disturbances $t^{1},t^{2},...,t^{G}$ ($G=8$) into the feature maps after the first CNN block of detector backbone and take the uncertainty of RPN classification feature maps as the uncertainty. Each pixel value of $t^{g}$ is sampled from a uniform distribution $U[-\gamma,+\gamma]$, where $\gamma = 0.3$. All SCL methods in our experiments are pretrained with the random sampling scheduler.

\subsection{The selection of loss and uncertainty for data difficulty measurer}

As for the two-stage ULD methods, there are at least four losses, i.e.,  RPN anchor classification loss and RPN box regression loss in Stage 1, and Region of Interest (RoI) box classification loss and regression loss in Stage 2. We need to select proper loss and feature maps (for uncertainty estimation)  to serve as difficulty measurements.

In this work, we adopt a `fixed one, test another' strategy to test the performance of two key components in two stages, i.e., RPN network in Stage 1 and ROI classification $\&$ regression in Stage 2. Specifically, we first train the network with its original network architectures and experiment settings to get a well-trained model weight. Then we replace the RoIs with GT Bounding Boxes (BBoxes) during the testing stage.  As shown in Tab. \ref{results}, we demonstrated this experiment in 2 SOTA ULD methods based on 50\% training data. Unsurprisingly, when the RoIs are fixed as the GT BBoxes, the performance suppresses the vanilla testing manner with a large margin and some of them are close to 100\%.

From this experiment, we can reach three conclusions that Stage 1 is more proper for data difficulty measurement. Hence we use the RPN anchor classification loss as the difficulty measure loss and take the RPN classification feature maps for uncertainty calculation  after adding disturbances into the feature maps after the first CNN block of the detector backbone.

\subsection{The relationship analysis between loss and uncertainty}
To show the necessity of introducing uncertainty, we proved that uncertainty and loss are two different dimensions to estimate the data difficulty in Lemma 1. Now we hereby give experimental proof to support it. Generally, we record all training samples loss and uncertainty of 2 SOTA ULD methods based on 25\%, 50\%, and 100\% training data, and draw their value or their indexes into 2D dashed figures. We hereby show the results based on \cite{li2022satr} in Fig. \ref{Fig3_loss_uncertainty}, other results are shown in the supplementary materials.

As shown in Fig. \ref{Fig3_loss_uncertainty} (a) and (b), all results show poor relativity between loss and uncertainty, hence Lemma 1 is confirmed. Specifically, we can find that the uncertainty is more sparse while loss is more concentrated (on small values). We believe the explanation for this is that the network training is a loss gradient descent process base on optimizers, which can directly decrease the training samples' loss but fail to directly solve the uncertainty problems.

Beyond Lemma 1, Fig. \ref{Fig3_loss_uncertainty} also shows some experimental results to support the superiority of our Mo-SCL method: i) When comparing the figure (a) and figure (b), we can find that after training with our Mo-SCL, the uncertainty becomes more concatenated (a2 v.s., b2) and shows the tendency of convergence (a1 v.s., b1). 2) As shown in (b), the index-based method can remove singular points in the value-based map (a), and the relationship between loss and uncertainty can be better illustrated. iii) Even though the network is converged based on a random sampling scheduler, Mo-SCL can further decrease the total loss and uncertainty stably over the whole training dataset. We believe that such a phenomenon can be explained with Corollary 4.2 that keeping  mini-batch's total difficulty even is helpful for the network's stable training.

\subsection{Lesion detection performance}
Two state-of-the-art ULD approaches \ucite{li2021conditional,li2022satr} are compared to evaluate MO-SCL's effectiveness.

\textbf{Partial training dataset results.} As shown in Table \ref{results}, under the $25\%$ and $50\%$ training data settings, the liner loss deweighting method is harmful to network training. Besides, the negative performance influences shrink with the number of training data increases. As for the anti-Mo methods, which pair low- (or high-) difficulty with low- (or high-) difficulty data, they can bring performance improvement for this mechanism will pair \{$<u^l,l^h>$,$<u^h,l^l>$\} together which both influences each other, but cause less effect on but decrease the effect on $<u^h,l^h>$ or $<u^l, l^l>$ data. This advantage will also shrink with the network generation increasing (i.e., train more training data).
The loss-based Mo-SCL methods bring improvement in the  $25\%$ training data setting but fail in $50\%$, the uncertainty-based Mo-SCL methods produce minor performance improvement in the $50\%$ training data setting. When combining them to form Mo-SCL produces the best result, which also shows the drawbacks of using single difficulty measurer methods discussed in Corollary 4.1.

\textbf{Full training dataset results.}  In Full dataset training, the proposed Mo-SCL follows a similar rule in Partial dataset training, but the improvements shrink with more training data avaialable. Please refer to the supplementary materials for more details.

\section{Conclusions}
In this paper, we propose Mo-SCL to simulate the fact that humans easily lose their concentration and interest if all the learning materials are relative easy (or hard).
The Mo-SCL combines a score by integrating both loss and uncertainty for better data difficulty estimation and proposes mixed-order sampling to alleviate the sample under-utilization problem. Theoretical investigations and extensive experiments based on ULD task show the superiority of our method, especially when the training data are scarce.

\newpage{}
{\small
\bibliographystyle{ieee_fullname}
\bibliography{egbib}

\begin{thebibliography}{10}

\bibitem{wang2021survey}
Xin Wang, Yudong Chen, and Wenwu Zhu.
\newblock A survey on curriculum learning.
\newblock {\em IEEE Trans. Pattern Anal. Mach. Intell.}, 2021.

\bibitem{kumar2010self}
M~Kumar, Benjamin Packer, and Daphne Koller.
\newblock Self-paced learning for latent variable models.
\newblock {\em Proc. Adv. Neural Inf. Process. Syst.}, 23, 2010.

\bibitem{jiang2014easy}
Lu~Jiang, Deyu Meng, Teruko Mitamura, and Alexander~G Hauptmann.
\newblock Easy samples first: Self-paced reranking for zero-example multimedia
  search.
\newblock In {\em ACMM}, pages 547--556, 2014.

\bibitem{zhao2015self}
Qian Zhao, Deyu Meng, Lu~Jiang, Qi~Xie, Zongben Xu, and Alexander~G Hauptmann.
\newblock Self-paced learning for matrix factorization.
\newblock In {\em IEEE AAAI}, 2015.

\bibitem{xu2015multi}
Chang Xu, Dacheng Tao, and Chao Xu.
\newblock Multi-view self-paced learning for clustering.
\newblock In {\em IJCAI}, 2015.

\bibitem{gong2018decomposition}
Maoguo Gong, Hao Li, Deyu Meng, Qiguang Miao, and Jia Liu.
\newblock Decomposition-based evolutionary multiobjective optimization to
  self-paced learning.
\newblock {\em IEEE Trans. on Evolut. Computa.}, 23(2):288--302, 2018.

\bibitem{bengio2009curriculum}
Yoshua Bengio, J{\'e}r{\^o}me Louradour, Ronan Collobert, and Jason Weston.
\newblock Curriculum learning.
\newblock In {\em ICML}, pages 41--48, 2009.

\bibitem{liu2022acpl}
Fengbei Liu, Yu~Tian, Yuanhong Chen, Yuyuan Liu, Vasileios Belagiannis, and
  Gustavo Carneiro.
\newblock Acpl: Anti-curriculum pseudo-labelling for semi-supervised medical
  image classification.
\newblock In {\em IEEE CVPR}, pages 20697--20706, 2022.

\bibitem{basu2022surpassing}
Soumen Basu, Mayank Gupta, Pratyaksha Rana, Pankaj Gupta, and Chetan Arora.
\newblock Surpassing the human accuracy: Detecting gallbladder cancer from usg
  images with curriculum learning.
\newblock In {\em IEEE CVPR}, pages 20886--20896, 2022.

\bibitem{roy2021curriculum}
Subhankar Roy, Evgeny Krivosheev, Zhun Zhong, Nicu Sebe, and Elisa Ricci.
\newblock Curriculum graph co-teaching for multi-target domain adaptation.
\newblock In {\em IEEE CVPR}, pages 5351--5360, 2021.

\bibitem{Huang_2020_CVPR}
Yuge Huang, Yuhan Wang, Ying Tai, Xiaoming Liu, Pengcheng Shen, Shaoxin Li,
  Jilin Li, and Feiyue Huang.
\newblock Curricularface: Adaptive curriculum learning loss for deep face
  recognition.
\newblock In {\em IEEE CVPR}, June 2020.

\bibitem{Pentina_2015_CVPR}
Anastasia Pentina, Viktoriia Sharmanska, and Christoph~H. Lampert.
\newblock Curriculum learning of multiple tasks.
\newblock In {\em IEEE CVPR}, June 2015.

\bibitem{meng2017theoretical}
Deyu Meng, Qian Zhao, and Lu~Jiang.
\newblock A theoretical understanding of self-paced learning.
\newblock {\em Information Sciences}, 414:319--328, 2017.

\bibitem{lee2011learning}
Yong~Jae Lee and Kristen Grauman.
\newblock Learning the easy things first: Self-paced visual category discovery.
\newblock In {\em CVPR}, pages 1721--1728. IEEE, 2011.

\bibitem{kumar2011learning}
M~Pawan Kumar, Haithem Turki, Dan Preston, and Daphne Koller.
\newblock Learning specific-class segmentation from diverse data.
\newblock In {\em ICCV}, pages 1800--1807. IEEE, 2011.

\bibitem{zhang2017spftn}
Dingwen Zhang, Le~Yang, Deyu Meng, Dong Xu, and Junwei Han.
\newblock Spftn: A self-paced fine-tuning network for segmenting objects in
  weakly labelled videos.
\newblock In {\em IEEE CVPR}, pages 4429--4437, 2017.

\bibitem{tang2012self}
Ye~Tang, Yu-Bin Yang, and Yang Gao.
\newblock Self-paced dictionary learning for image classification.
\newblock In {\em ACMM}, pages 833--836, 2012.

\bibitem{Ghasedi_2019_CVPR}
Kamran Ghasedi, Xiaoqian Wang, Cheng Deng, and Heng Huang.
\newblock Balanced self-paced learning for generative adversarial clustering
  network.
\newblock In {\em IEEE CVPR}, June 2019.

\bibitem{III_2013_CVPR}
James~S. Supancic, III and Deva Ramanan.
\newblock Self-paced learning for long-term tracking.
\newblock In {\em IEEE CVPR}, June 2013.

\bibitem{tang2012shifting}
Kevin Tang, Vignesh Ramanathan, Li~Fei-Fei, and Daphne Koller.
\newblock Shifting weights: Adapting object detectors from image to video.
\newblock {\em Proc. Adv. Neural Inf. Process. Syst.}, 25, 2012.

\bibitem{zhang2019leveraging}
Dingwen Zhang, Junwei Han, Long Zhao, and Deyu Meng.
\newblock Leveraging prior-knowledge for weakly supervised object detection
  under a collaborative self-paced curriculum learning framework.
\newblock {\em IEEE IJCV}, 127(4):363--380, 2019.

\bibitem{zhou2018deep}
Sanping Zhou, Jinjun Wang, Deyu Meng, Xiaomeng Xin, Yubing Li, Yihong Gong, and
  Nanning Zheng.
\newblock Deep self-paced learning for person re-identification.
\newblock {\em Pattern Recognition}, 76:739--751, 2018.

\bibitem{han2019weakly}
Junwei Han, Yang Yang, Dingwen Zhang, Dong Huang, Dong Xu, and Fernando
  De~La~Torre.
\newblock Weakly-supervised learning of category-specific 3d object shapes.
\newblock {\em IEEE Trans. Pattern Anal. Mach. Intell.}, 43(4):1423--1437,
  2019.

\bibitem{ghasedi2019balanced}
Kamran Ghasedi, Xiaoqian Wang, Cheng Deng, and Heng Huang.
\newblock Balanced self-paced learning for generative adversarial clustering
  network.
\newblock In {\em IEEE CVPR}, pages 4391--4400, 2019.

\bibitem{gong2016multi}
Chen Gong, Dacheng Tao, Stephen~J Maybank, Wei Liu, Guoliang Kang, and Jie
  Yang.
\newblock Multi-modal curriculum learning for semi-supervised image
  classification.
\newblock {\em IEEE Trans. on Imag. Proces.}, 25(7):3249--3260, 2016.

\bibitem{zhang2015self}
Dingwen Zhang, Deyu Meng, Chao Li, Lu~Jiang, Qian Zhao, and Junwei Han.
\newblock A self-paced multiple-instance learning framework for co-saliency
  detection.
\newblock In {\em IEEE CVPR}, pages 594--602, 2015.

\bibitem{li2017self}
Changsheng Li, Fan Wei, Junchi Yan, Xiaoyu Zhang, Qingshan Liu, and Hongyuan
  Zha.
\newblock A self-paced regularization framework for multilabel learning.
\newblock {\em IEEE Trans. Neural Netw. Learn Syst.}, 29(6):2660--2666, 2017.

\bibitem{li2017self2}
Changsheng Li, Junchi Yan, Fan Wei, Weishan Dong, Qingshan Liu, and Hongyuan
  Zha.
\newblock Self-paced multi-task learning.
\newblock In {\em IEEE AAAI}, 2017.

\bibitem{lin2017active}
Liang Lin, Keze Wang, Deyu Meng, Wangmeng Zuo, and Lei Zhang.
\newblock Active self-paced learning for cost-effective and progressive face
  identification.
\newblock {\em IEEE Trans. Pattern Anal. Mach. Intell.}, 40(1):7--19, 2017.

\bibitem{tang2019self}
Ying-Peng Tang and Sheng-Jun Huang.
\newblock Self-paced active learning: Query the right thing at the right time.
\newblock In {\em IEEE AAAI}, volume~33, pages 5117--5124, 2019.

\bibitem{mackay1992practical}
D.~MacKay.
\newblock A practical bayesian framework for backpropagation networks.
\newblock {\em Neural Comput}, 4(3):448--472, 1992.

\bibitem{gal2016dropout}
Y.~Gal and Z.~Ghahramani.
\newblock Dropout as a bayesian approximation: Representing model uncertainty
  in deep learning.
\newblock In {\em ICML}, pages 1050--1059. PMLR, 2016.

\bibitem{teye2018bayesian}
M.~Teye, H.~Azizpour, and K.~Smith.
\newblock Bayesian uncertainty estimation for batch normalized deep networks.
\newblock In {\em ICML}, pages 4907--4916. PMLR, 2018.

\bibitem{lakshminarayanan2016simple}
B.~Lakshminarayanan, A.~Pritzel, and C.~Blundell.
\newblock Simple and scalable predictive uncertainty estimation using deep
  ensembles.
\newblock {\em arXiv preprint arXiv:1612.01474}, 2016.

\bibitem{yu2019uncertainty}
L.~{Yu}, S.~{Wang}, X.~{Li}, C.~{Fu}, and P.~{Heng}.
\newblock Uncertainty-aware self-ensembling model for semi-supervised 3d left
  atrium segmentation.
\newblock In {\em MICCAI}, pages 605--613. Springer, 2019.

\bibitem{luo2020deep}
L.~{Luo} et~al.
\newblock Deep mining external imperfect data for chest x-ray disease
  screening.
\newblock {\em IEEE Trans Med Imaging}, 39(11):3583--3594, 2020.

\bibitem{xia2020uncertainty}
Y.~{Xia} et~al.
\newblock Uncertainty-aware multi-view co-training for semi-supervised medical
  image segmentation and domain adaptation.
\newblock {\em Med. Image Anal.}, 65:101766, 2020.

\bibitem{mehrtash2020confidence}
A.~Mehrtash, W.~M. Wells, C.~M. Tempany, P.~Abolmaesumi, and T.~Kapur.
\newblock Confidence calibration and predictive uncertainty estimation for deep
  medical image segmentation.
\newblock {\em IEEE Trans Med Imaging}, 39(12):3868--3878, 2020.

\bibitem{shi2021inconsistency}
Y.~{Shi} et~al.
\newblock Inconsistency-aware uncertainty estimation for semi-supervised
  medical image segmentation.
\newblock {\em IEEE Trans Med Imaging}, 2021.

\bibitem{shrivastava2016training}
Abhinav Shrivastava, Abhinav Gupta, and Ross Girshick.
\newblock Training region-based object detectors with online hard example
  mining.
\newblock In {\em IEEE CVPR}, pages 761--769, 2016.

\bibitem{zhang2019anchor_free}
N.~Zhang~et al.
\newblock 3d anchor-free lesion detector on computed tomography scans.
\newblock {\em arXiv:1908.11324}, 2019.

\bibitem{zhang2020Agg_Fas}
N.~Zhang~et al.
\newblock 3d aggregated faster {R-CNN} for general lesion detection.
\newblock {\em arXiv:2001.11071}, 2020.

\bibitem{yan20183DCE}
K.~Yan~et al.
\newblock 3d context enhanced region-based convolutional neural network for
  end-to-end lesion detection.
\newblock In {\em MICCAI}, pages 511--519. Springer, 2018.

\bibitem{li2019mvp}
Z.~Li~et al.
\newblock Mvp-net: Multi-view fpn with position-aware attention for deep
  universal lesion detection.
\newblock In {\em MICCAI}, pages 13--21. Springer, 2019.

\bibitem{yan2019mulan}
K.~Yan~et al.
\newblock {Mulan}: Multitask universal lesion analysis network for joint lesion
  detection, tagging, and segmentation.
\newblock In {\em MICCAI}, pages 194--202. Springer, 2019.

\bibitem{yang2020alignshift}
J.~Yang~et al.
\newblock Alignshift: bridging the gap of imaging thickness in 3d anisotropic
  volumes.
\newblock In {\em MICCAI}, pages 562--572. Springer, 2020.

\bibitem{cai2020deep}
J.~Cai~et al.
\newblock Deep volumetric universal lesion detection using light-weight pseudo
  3d convolution and surface point regression.
\newblock In {\em MICCAI}, pages 3--13. Springer, 2020.

\bibitem{zhang2020revisiting}
S.~Zhang~et al.
\newblock Revisiting 3d context modeling with supervised pre-training for
  universal lesion detection in ct slices.
\newblock In {\em MICCAI}, pages 542--551. Springer, 2020.

\bibitem{tang2021weakly}
Y.~Tang~et al.
\newblock Weakly-supervised universal lesion segmentation with regional level
  set loss.
\newblock In {\em MICCAI}, pages 515--525. Springer, 2021.

\bibitem{yang2021asymmetric}
J.~Yang~et al.
\newblock Asymmetric 3d context fusion for universal lesion detection.
\newblock In {\em MICCAI}, pages 571--580. Springer, 2021.

\bibitem{li2021conditional}
H.~Li~et al.
\newblock Conditional training with bounding map for universal lesion
  detection.
\newblock In {\em MICCAI}, pages 141--152. Springer, 2021.

\bibitem{lyu2021segmentation}
F.~Lyu~et al.
\newblock A segmentation-assisted model for universal lesion detection with
  partial labels.
\newblock In {\em MICCAI}, pages 117--127. Springer, 2021.

\bibitem{ouali2020semi}
Y.~{Ouali}, C.~{Hudelot}, and M.~{Tami}.
\newblock Semi-supervised semantic segmentation with cross-consistency
  training.
\newblock In {\em CVPR}, pages 12674--12684, 2020.

\bibitem{li2022satr}
Han Li, Long Chen, Hu~Han, and S~Kevin~Zhou.
\newblock Satr: Slice attention with transformer for universal lesion
  detection.
\newblock In {\em MICCAI}, pages 163--174. Springer, 2022.

\bibitem{yan18deeplesion}
K.~Yan~et al.
\newblock Deep lesion graphs in the wild: relationship learning and
  organization of significant radiology image findings in a diverse large-scale
  lesion database.
\newblock In {\em IEEE CVPR}, pages 9261--9270, 2018.

\end{thebibliography}
}

\end{document}